\documentclass[letterpaper, 10 pt, conference]{ieeeconf}  

\IEEEoverridecommandlockouts                              

\usepackage{standalone}
\usepackage[textsize=tiny]{todonotes}
\usepackage{tabularx}
\usepackage{pgfplots, pgfplotstable}
\usepackage{textcomp}
\usepackage{makecell}
\usepackage{booktabs}
\usepackage{listings}
\usepackage{lipsum}
\usepackage{multirow}
\usepackage{color}
\usepackage[T1]{fontenc}
\usepackage{amsmath, amssymb}
\usepackage{graphicx}
\usepackage{subcaption}
\usepackage[labelfont=bf]{caption}
\usepackage{bm}
\usepackage{afterpage}

\usepackage{amsmath}        
\usepackage[ruled, vlined, linesnumbered]{algorithm2e}

\usepackage{float}

\makeatletter
\newcommand\BeraMonottfamily{%
  \def\fvm@Scale{0.85}
  \fontfamily{fvm}\selectfont
}
\makeatother

\usepackage{xcolor}
\usepackage{array}
\usepackage{wasysym}
\usepackage[flushleft]{threeparttable}

\usepackage[T1]{fontenc}
\usepackage{preamble}
\usepackage{cleveref}
\usepackage{graphicx}
\usepackage{booktabs}
\usepackage{pifont}

\newcommand{\uniformtablesize}{%
  \scriptsize
  \setlength{\tabcolsep}{6pt}
  \renewcommand{\arraystretch}{0.9}
}

\title{\LARGE \bf

SaferPath: Hierarchical Visual Navigation with Learned Guidance and Safety-Constrained Control

}

\author{Lingjie Zhang, Zeyu Jiang, Changhao Chen$^{*}$
\thanks{Lingjie Zhang, Zeyu Jiang and Changhao Chen are with PEAK-Lab, The Hong Kong University of Science and Technology (Guangzhou), Guangzhou, 511453, China
        ({\tt\small zljjacob@gmail.com, zjiang122@connect.hkust-gz.edu.cn, changhaochen@hkust-gz.edu.cn})}%
\thanks{This work was supported by National Natural Science Foundation of China (NFSC) under the Grant Number 62573370 and Key Area Project of Education Department of Guangdong Province (No. 2025ZDZX3051).}
\thanks{$^*$ Corresponding Author. }
}

\begin{document}

\newcommand{\jh}[1]{{ \color{blue}[jh: #1]}}

\maketitle
\thispagestyle{empty}
\pagestyle{empty}

\begin{abstract}
  Visual navigation is a core capability for mobile robots, yet end-to-end learning-based methods often struggle with generalization and safety in unseen, cluttered, or narrow environments. These limitations are especially pronounced in dense indoor settings, where collisions are likely and end-to-end models frequently fail. To address this, we propose SaferPath, a hierarchical visual navigation framework that leverages learned guidance from existing end-to-end models and refines it through a safety-constrained optimization-control module. SaferPath transforms visual observations into a traversable-area map and refines guidance trajectories using Model Predictive Stein Variational Evolution Strategy (MP-SVES), efficiently generating safe trajectories in only a few iterations. The refined trajectories are tracked by an MPC controller, ensuring robust navigation in complex environments. Extensive experiments in scenarios with unseen obstacles, dense unstructured spaces, and narrow corridors demonstrate that SaferPath consistently improves success rates and reduces collisions, outperforming representative baselines such as ViNT and NoMaD, and enabling safe navigation in challenging real-world settings.
\end{abstract}

\section{Introduction}
Navigation is a fundamental capability of mobile robots, enabling autonomous operation and the execution of complex tasks in dynamic and uncertain environments. Due to limited prior knowledge and the inherent unpredictability of the real world, enhancing navigation in previously unseen environments remains one of the central challenges in robotics \cite{liu2025embodied, ZHANG2022105036}.
Traditional navigation approaches typically rely on pre-built maps (e.g., occupancy grids \cite{elfes1989occupancygrids} or topological graphs \cite{thrun1996integrating}) to plan paths. These methods often require complex sensor fusion, such as integrating LiDAR or depth cameras, and planning algorithms such as A* \cite{Hart1968heuristicdetermination} or RRT \cite{LaValle1998RapidlyexploringRT}.


With the rapid advancement of embodied AI, end-to-end visual navigation has attracted growing attention. These methods leverage deep neural networks (DNNs) to directly map visual inputs to actions, reducing reliance on explicit environmental modeling and planning \cite{liu2025embodied}. However, their performance depends heavily on large-scale, diverse training datasets for robust generalization. Collecting and annotating such datasets is costly, and distribution mismatches between training and deployment environments can lead to significant performance degradation. Furthermore, real-world scenarios often include unseen obstacles, unstructured layouts, and narrow passages—for example, indoor navigation for domestic robots—where collisions are likely and safety is critical. Diffusion-based navigation models \cite{sridhar2023nomad} are particularly prone to performance degradation in these settings, as confirmed in our experiments.

\begin{figure}[t]
    \centering
    \captionsetup{font=small}
    \includegraphics[width=\linewidth]{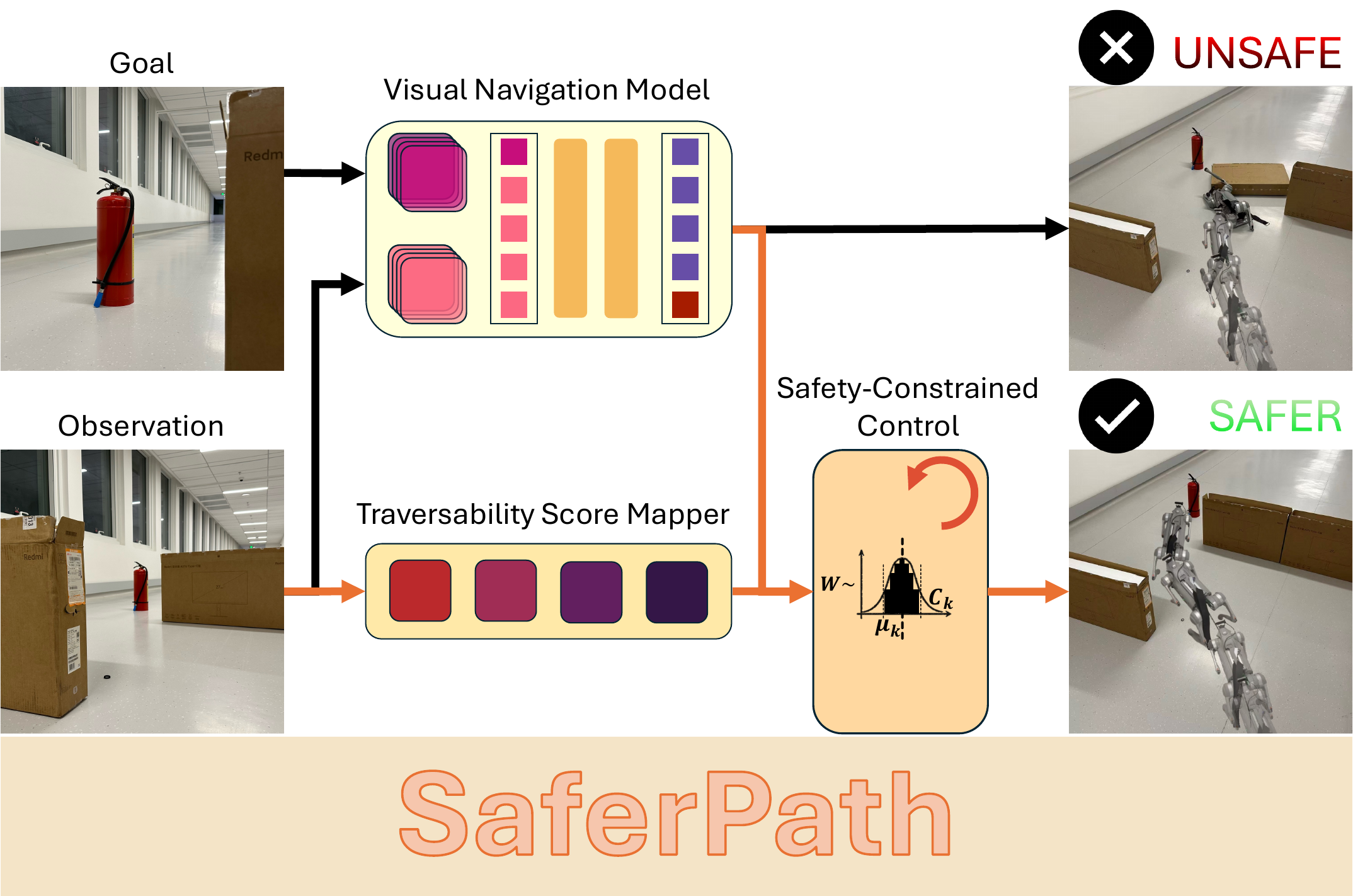}
    \caption{SaferPath is a hierarchical navigation framework that uses a end-to-end visual navigation model for guidance and integrates safety-constrained control module, significantly improving obstacle avoidance and narrow-passage traversal for safe navigation in challenging environments.}
    \label{fig:teaser}
    \vspace{-20pt}
\end{figure}

This raises two fundamental questions:

\begin{enumerate}
\item[\textbf{Q1.}] How can the safety and robustness of end-to-end visual navigation models be improved without retraining when deployed in diverse and dynamic environments?
\item[\textbf{Q2.}] Since training data often lack dense and narrow scenarios, can the proposed approach enhance navigation performance and safety in such challenging environments?
\end{enumerate}

\begin{figure*}[htbp]
    \centering
    \vspace{5pt}
    \captionsetup{font=small}
    \includegraphics[width=1\linewidth]{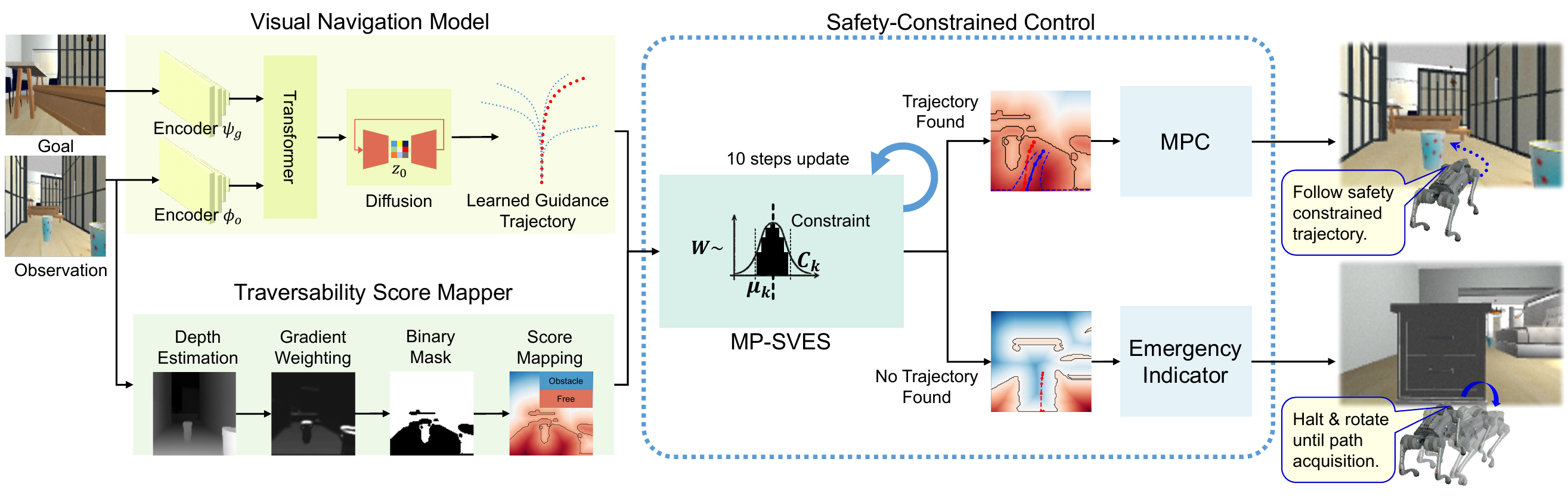}
    \caption{SaferPath Pipeline overview. SaferPath can be integrated with various end-to-end visual navigation models. The Traversability Score Mapper processes RGB observations to generate a score map $\mathcal{S}$, which, together with the learned guidance trajectory, is passed to the MP-SVES module. If optimization fails, the Emergency Indicator stops the robot and rotates it until a traversable trajectory is found; otherwise, the MPC controller executes trajectory tracking.}
    \label{fig:framework}
    \vspace{-15pt}
\end{figure*}

To address these challenges, we propose SaferPath, a novel hierarchical visual navigation framework with learned guidance and safety-contrained control. SaferPath is designed as a flexible modular layer above end-to-end navigation models. Instead of directly executing the trajectories predicted by an end-to-end model, we use them only as guidance trajectories. Specifically, visual observations are first transformed into a traversable-area map. The learned guidance trajectory is then refined by formulating the problem as a model predictive optimization task. To reduce the high computation requirements of nonlinear optimization, we introduce the Model Predictive Stein Variational Evolution Strategy (MP-SVES), which recasts trajectory refinement as an efficient variational inference process. MP-SVES generates safety-constrained solutions within only a few iterations, thereby avoiding the heavy computational cost of conventional nonlinear solvers. The refined, safety-constrained trajectory is subsequently tracked by an MPC controller, ensuring robust and precise navigation in complex environments. 

Across diverse scenarios, SaferPath consistently outperforms strong baselines, maintaining high success rates and low collision counts even in cluttered or narrow environments. Compared to the representative visual navigation model NoMaD \cite{sridhar2023nomad}, SaferPath improves performance by 40\% in robust navigation under unseen obstacles and over 50\% in exploration within dense unstructured environments, while successfully navigating narrow corridors where all baselines fail.

In summary, our main contributions are as follows.
\begin{itemize}
    \item We propose SaferPath, a hierarchical visual navigation framework that flexibly builds on end-to-end models by leveraging learned guidance while incorporating safety-constrained control, thereby enhancing navigation accuracy and robustness in complex unknown environments.
    \item We introduce Model Predictive Stein Variational Evolution Strategy (MP-SVES), a novel optimization algorithm that efficiently solves nonconvex nonlinear problems and generates feasible, safety-constrained trajectories within only a few iterations.
    \item We conduct extensive experiments across diverse scenarios—including navigation with unseen obstacles, exploration in dense unstructured environments, and obstacle avoidance in narrow indoor passages—demonstrating that SaferPath significantly outperforms representative baselines such as ViNT and NoMaD.
\end{itemize}




\section{Related Work}

\subsection{Embodied Navigation}

With the development of embodied intelligence, learning-based methods, in contrast to geometry-based approaches, emphasize the development of reliable navigation strategies by learning from data-driven navigation experiences\cite{fa17a395a25b40f7839626358d6968b3}. 

Typically, learning-based navigation strategies are formulated as sequence prediction tasks, where current and historical observations are exploited to generate future behaviors under a generative modeling paradigm\cite{zheng2024genad, hu2023_uniad, badgr}. Several studies have explored end-to-end navigation frameworks that integrate heterogeneous sensing modalities, including LiDAR, depth perception, and proprioceptive measurements, to enhance long-range global navigation performance \cite{liang2024mtg, liang2024dtg, saviolo2025novanavigationobjectcentricvisual}. ViNT\cite{shah2023vint} and other recent works\cite{sridhar2023nomad, shah2022gnm, zeng2025navidiffusor} propose purely vision-based, end-to-end, map-free robot navigation approaches, where models directly process sequences of visual observations to infer future motions or navigation actions, enabling autonomous navigation in complex or previously unseen environments.

\subsection{Learning-Enhanced Model Predictive Control}

Model Predictive Control (MPC) uses known system dynamics to optimize control inputs for predicting and regulating future states. Its strengths in real-time optimization and constraint handling make it ideal for motion planning in embodied intelligent robots\cite{Grandia2019frequencyaware}. However, MPC is less effective for long-horizon predictions due to increasing computational complexity and accumulation of model inaccuracies over time.

Recent works have integrated model-based control approaches with generative models, combining the flexibility of these models with the real-time performance and robustness of MPC to enhance control accuracy and efficiency in robotic systems. Huang et al. introduced the Subgoal Diffuser, a diffusion-based framework that generates adaptive, coarse-to-fine subgoals to guide MPC, enabling robust planning for long-horizon robotic manipulation tasks\cite{huang2024subgoaldiffuser}. In SICNav-Diffusion, the diffusion model is integrated with the MPC framework to facilitate safe trajectory prediction in crowd navigation environments\cite{samavi2025sicnav}. Other researchers\cite{xue2024fullordersamplingbasedmpctorquelevel} also try to combine MPC with a generative model. 
In this work, we integrate end-to-end visual navigation models with MPC-based prediction under safety constraints, addressing the limitations of purely end-to-end approaches and achieving more reliable and robust navigation in real-world environments.

\section{Method}

Prior end-to-end navigation models often improve performance via sophisticated designs or larger datasets, yet still generalize poorly to unseen environments. 
To address this, we propose a hierarchical navigation framework where a end-to-end visual navigation model providing learned guidance trajectory based on current and past observations while a seperate optimization and control module enforces safety and dynamic constraints. This design reduces reliance on the end-to-end model, requiring it only to provide coarse directional guidance for more reliable navigation. Figure \ref{fig:framework} illustrates the overall hierarchical navigation framework, with further details provided in the following sections.

\subsection{Preliminaries}
SaferPath builds upon end-to-end visual navigation models. These models take the robot’s current and historical RGB observations as input, optionally conditioned on a target image or target position. The inputs are first processed by a visual encoder to produce latent representations, which are then passed through a downstream neural network to predict the trajectory. Formally, the model can be expressed as:
\begin{equation}
f_\theta: \mathbb{R}^{L\times H\times W\times C}\times \mathcal{G}\rightarrow \mathbf{X}_t,\quad\mathbf{X}_t=f_\theta(\mathbf{I}_t,g),
\end{equation}
where $\mathbf{I}_t=\{I_{t-L+1},\dots,I_t\},\;\mathbf{I}_t\in\mathbb{R}^{L\times H\times W\times C}$ denotes the current and historical observations, where $L$ is the history length, $H$ and $W$ denote the image height and width, and $C$ is the number of channels. $g \in \mathcal{G}$ is the goal image, which is typically one of a sequence of images organized as a topological map along the navigation path.  $\mathbf{X}_t=\{\mathbf{x}_0,\dots,\mathbf{x}_{t+T}\}$ is the predicted trajectory with $\mathbf{X}_t\in\mathbb{R}^d$ representing the robot's position trajectory over time. 

For instance, we use NoMaD \cite{sridhar2023nomad} as a base end-to-end model. NoMaD employs an EfficientNet-B0 encoder \cite{tan2020efficientnetrethinkingmodelscaling} to encode both the observation sequence $\mathbf{I}_t$ and the goal image $g$, producing tokenized latent representations $z_t$. These tokens are then processed by Transformer layers \cite{vaswani2023attentionneed} to generate a contextual vector $c_t$, which a diffusion model \cite{ho2020denoisingdiffusionprobabilisticmodels} uses to predict the trajectory $\mathbf{X}_t$. Additionally, NoMaD supports a goal-mask input, enabling exploration without an explicit goal.


\subsection{Traversability Score Mapper}

The Traversability Score Mapper (TSM) module constructs a safety-aware score map from RGB images to delineate navigable regions for trajectory planning. First, depth estimation is performed using the DepthAnythingV2 model \cite{depthanything}, producing a normalized depth map in the range $[0, 1]$. Gaussian smoothing reduces noise, followed by vertical gradient computation via a Sobel operator to highlight depth discontinuities associated with obstacles. To emphasize regions near the robot, a spatially varying weight that increases from top to bottom of the image is applied. The weighted gradient is thresholded to generate a preliminary binary map, labeling obstacles as 0 and traversable areas as 1, with adjustments near the bottom rows to prevent misclassification.

This binary map is refined using horizontal and vertical gradient analysis to detect boundaries and segment obstacle versus free-space regions. A Euclidean distance transform then computes pixel-wise distances to these regions, yielding a smooth score map. In this map, costs are defined as: positive near obstacles (increasing with distance from boundaries), negative in free-space (decreasing with distance), and zero at boundary transitions. Finally, the map is normalized to $[-1, 1]$, producing a continuous score map $\mathcal{S}$ that encapsulates safe navigation regions for trajectory planning.


\subsection{Safety-Constrained Control}

We employ the score map $\mathcal{S}$ from TSM as a safety constraint for trajectory planning. Since $\mathcal{S}$ is nonconvex, nonlinear, and highly variable, solving the optimal control problem directly using classic methods such as sequential quadratic programming (SQP)\cite{bonnans2006numerical} or interior point methods (IPM)\cite{pdipm} is computationally expensive and unsuitable for real-time operation. To overcome this, we propose \textbf{M}odel \textbf{P}redictive \textbf{S}tein \textbf{V}ariational \textbf{E}volution \textbf{S}trategy (MP-SVES), which reformulates trajectory optimization as a model predictive task. MP-SVES combines sampling-based methods \cite{theodorou10a, mppi2016} with Stein variational strategies \cite{liu2019steinvariationalgradientdescent, Braun2024SteinVE} to accelerate convergence. The safety constrained trajectory is then executed using an MPC control module.

\begin{algorithm}[t]
\caption{Trajectory Optimization via MP-SVES}
\label{alg:traj_opt}
\KwIn{Learned guidance trajectory $\mathbf{X}_{\text{ref},t}$, score map $\mathcal{S}$}
\KwOut{Safety constrained trajectory $\mathbf{X}_{\text{opt},t}$}
\SetAlgoLined
\textbf{Parameters:} Particle number $P$, sample number $m$, iteration number $N$, kernel weight $\gamma$, step size $\eta$\\
Initialize $\bm{\mu}_p$, $\mathbf{C}_p$ for $p = 1, \dots, P$\\
\For{each timestep $t$}{
    \For{$n = 1$ \KwTo $N$}{
        \For{$p = 1$ \KwTo $P$}{
            $\bm\epsilon_{p,i}\sim\mathcal{N}(\mathbf{0},\mathbf{I}),\;\text{for}\;i=1,\dots,m$\\
            $\mathbf{W}_{p,i} \gets \bm\mu_p+\mathbf{L_p}\bm\epsilon_{p,i},\;\mathbf{C}_p=\mathbf{L_p}\mathbf{L_p}^{\top}$\\
            $J_{p,i} \gets J(\mathbf{W}_{p,i}, \mathcal{S}, \mathbf{X}_{\text{ref}})$\\
            $\omega_{p,i} \gets \frac{\exp(-J_{p,i})}{\sum_{j=1}^m\exp(-J_{p,j})}$\\
            $\Delta\mathbf{\bar{W}}_p \gets \sum_{i=1}^m \omega_{p,i} (\mathbf{W}_{p,i}-\bm\mu_p)$\\
            $\mathbf{g}^{\text{ker}}_p \gets \sum_{q=1 , q \neq p}^{P-1}\nabla_{\bm{\mu_p}}{\text{ker}}(\bm\mu_p, \bm\mu_q)$\\
            $\bm\mu_p \leftarrow \bm\mu_p + \eta \cdot (\Delta\mathbf{\bar{W}}_p + \gamma \mathbf{g}^{\text{ker}}_p)$\\
            $\mathbf{C}_p \leftarrow \mathbf{C}_p + \eta \cdot \sum_{i=1}^{m}\omega_{p,i} \bm\epsilon_{p,i}\bm\epsilon^{\top}_{p,i}$\\
        }
    }
    $\mathbf{W}^*_t \gets \arg\underset{\mathbf{W}_{p,i}}{\min} \, J_{p,i} \;\; \text{s.t. constraints}$\\
    $\mathbf{X}_{\text{opt},t} \gets \mathbf{W}_t^*\cdot \mathbf{X}_{\text{ref},t}$\\
}
\end{algorithm}

\textbf{MP-SVES Optimization:} In the first step, the learned guidance trajectory $\mathbf{X}_{\text{ref}} = \left\{ \begin{bmatrix} x_{\text{ref}}^{(k)} , y_{\text{ref}}^{(k)} \end{bmatrix}^\top \right\}_{k=0}^{K-1}$ from end-to-end visual navigation model is passed to the MP-SVES optimization module. The goal of this module is to optimize the learned guidance trajectory by adjusting its parameters while ensuring that the safety constrained trajectory remains as close as possible to learned guidance trajectory, while also satisfying safety constraints defined by the score map $\mathcal{S}$. The optimization is formulated by applying an anisotropic scaling to the learned guidance trajectory points:
\begin{equation}
\mathbf{X}_{\text{opt}}^{(k)} = \begin{bmatrix} x_{\text{opt}}^{(k)} \\ y_{\text{opt}}^{(k)} \end{bmatrix} = \begin{bmatrix} w_x & 0 \\ 0 & w_y \end{bmatrix} \begin{bmatrix} x_{\text{ref}}^{(k)} \\ y_{\text{ref}}^{(k)} \end{bmatrix}, \quad k = 0, \dots, K-1,
\end{equation}
where $w_x$ and $w_y$ are optimization parameters controlling the scaling of learned guidance trajectory in the $x$ and $y$ dimensions.
This scaling-based approach allows the optimization to preserve key geometric characteristics of the original trajectory, such as general shape and heading, while providing flexibility to adjust its size and improve compliance with safety constraints.

The MP-SVES optimization algorithm formulates trajectory optimization as a model predictive problem, where the goal is to identify the optimal scaling parameters $\mathbf{W} = \mathrm{diag}(w_x, w_y)$ that minimize the following objective function subject to system dynamics and feasibility constraints:

\vspace{-15pt}
\begin{equation}
\begin{aligned}
\label{eq:mp_sves}
\min _{\mathbf{W}}J(\mathbf{W}, \mathcal{S}, \mathbf{X}_{\text{ref}}) &= \sum_{k=0}^{K-1} \| \mathbf{X}_{\text{opt}}^{(k)} - \mathbf{X}_{\text{ref}}^{(k)} \|_Q^2 \\
&\quad + \lambda_s \sum_{k=0}^{K-1} \mathcal{S}\!\left(\mathcal{X}_{\text{tube}}^{(k)}\right) + \lambda_r \| \mathbf{W} - \mathbf{I} \|_2^2\\
\text{s. t.} \quad &\mathbf{X}_{\text{opt, init}} = \mathbf{X}_{\text{ref}}\\
& \mathbf{X}_{\text{opt}}^{(k)} = \mathbf{W}\cdot\mathbf{X}_{\text{ref}}^{(k)}, \quad k = 0,\dots, K-1\\
& \mathcal{S}\!\left(\mathcal{X}_{\text{tube}}^{(k)}\right) \leq \delta, \quad k = 0, \dots, K-1\\
\text{where} \quad &
\mathcal{X}_{\text{tube}}^{(k)} =
\left\{ \mathbf{X}_{\text{opt}}^{(k)} + \alpha \,\mathbf{n}^{(k)} \;\big|\; \alpha \in [\frac{-d}{2}, \frac{d}{2}] \right\},
\end{aligned}
\end{equation}



$K$ is the prediction horizon at each timestep. $Q$, $\lambda_s$, and $\lambda_r$ are tuning parameters controlling tracking accuracy, safety, and regularization. $d$ denotes the robot width.$n^{(k)}$ denotes the unit normal vector orthogonal to the trajectory at timestep $k$.

This formulation seeks scaling parameters that keep safety constrained trajectory close to the learned guidance trajectory while satisfying the safety constraints. 
The MP-SVES iterative procedure used to update $\mathbf W$ and obtain safety constrained trajectory is summarized in Algorithm~\ref{alg:traj_opt}, where $\bm{\mu_p}$ and $\mathbf{C_p}$ represent the mean and covariance matrix of particle $p$. The repulsion gradient, ${\nabla_{\mu_p} \text{ker}(\bm\mu_p, \bm\mu_q)}$, is the gradient of the kernel function, typically the RBF kernel $\text{ker}(\bm\mu_p,\bm\mu_q) = \exp\left(-\frac{\|\bm\mu_p - \bm\mu_q\|^2}{2\sigma^2}\right)$, generating repulsive forces to prevent mode collapse and maintain diversity. ${\mathbf{W}_t^*}$ denotes the optimal scaling parameters that satisfy safety constraints.

\begin{figure}[t]
    \centering
    \vspace{5pt}
    \captionsetup{font=small}
    \includegraphics[width=\linewidth]{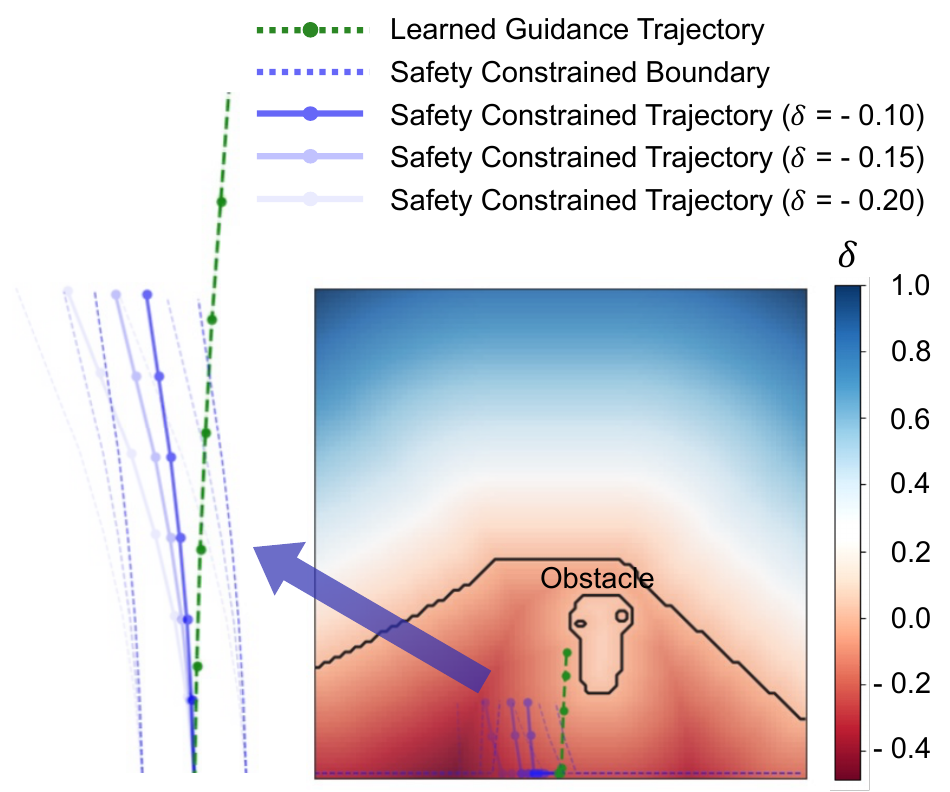}
    \caption{Safety constrained trajectories under different safety thresholds, with the robot maintaining varying safe distances from obstacles by adjusting threshold $\delta$}
    \label{fig:svpi_success}
    \vspace{-10pt}
\end{figure}

In practice, MP-SVES algorithm requires only about 10 iterations to obtain a feasible safety constrained trajectory, making it significantly more efficient than gradient descent methods that directly optimize over the score map. When facing obstacles, we vary the safety threshold to show how safety constrained trajectories change: Smaller (more negative) thresholds lead to stricter safety constraints, causing safety constrained trajectories to deviate further from obstacles (Fig.~\ref{fig:svpi_success}). In cases where obstacles are too close and no feasible solution can be found, the situation is classified as dangerous. In such cases, the Emergency Indicator commands the robot to stop immediately and rotate toward the direction with a larger free-space region until a feasible safety constrained trajectory is obtained.

\textbf{Model Predictive Control:} Once a safety constrained trajectory has been obtained, it is passed to the MPC module. The purpose of this module is to compute the optimal control inputs $ \mathbf{U}_t = (v_t, \omega_t)$ at each time step $t$, ensuring that the robot's state $\mathbf{X}_t = (x_t, y_t, \theta_t)$ closely follows safety constrained trajectory $\mathbf{X}_{\text{opt}}$, while respecting system dynamics, physical constraints. The MPC problem is formulated as:
\vspace{-10pt}

\begin{equation}
\begin{aligned}
&\min_{\mathbf{U}_t^{(0)}, \dots, \mathbf{U}_{t}^{(K-1)}} 
\Bigg\{ 
\sum_{k=0}^{K-1} 
\Big( 
   \| \mathbf{X}_{t}^{(k)} - \mathbf{X}_{\text{opt},\,t}^{(k)} \|_{Q}^{2} 
   + \| \mathbf{U}_{t}^{(k)} \|_{R}^{2} 
\Big) \\[4pt]
&\hspace{5em}
+ \underbrace{\| \mathbf{X}_{t}^{(K)} - \mathbf{X}_{\text{opt},\,t}^{(K)} \|_{Q_f}^{2}}_{\text{terminal cost}}
\Bigg\} \\[6pt]
&\text{s.t.}\quad
\begin{aligned}[t]
\mathbf{X}_t^{(0)} &= \mathbf{X}_{\text{opt},t}^{(0)},\\
\mathbf{X}_{t}^{(k+1)} &= f(\mathbf{X}_{t}^{(k)}, \mathbf{U}_{t}^{(k)}),\quad k=0,\dots,K-1,\\
\mathbf{X}_{t}^{(k)} &\in \mathcal{X},\;\mathbf{U}_{t}^{(k)} \in \mathcal{U},\quad k=0,\dots,K-1,\\
\end{aligned}
\end{aligned}
\end{equation}

where $f(\cdot)$ denotes the robot's kinematic model, $ \mathbf{U}_{t}^{(k)} = (v_{t}^{(k)}, \omega_{t}^{(k)}) $ is the control input at time $ t+k $, $ \mathcal{X} $ and $ \mathcal{U} $ are the feasible state and control input constraint sets.

The objective of the MPC is to penalize deviations from safety constrained trajectory $\mathbf{X}_{\text{opt}}$ and excessive control inputs, while ensuring that the trajectory remains within the dynamic constraints. 

The robot's kinematic model $f(\cdot)$ is defined as:

\begin{equation}
\begin{aligned}
x_{t+1} &= x_t + v_t \cos(\theta_t) \Delta t, \\
y_{t+1} &= y_t + v_t \sin(\theta_t) \Delta t, \\
\theta_{t+1} &= \theta_t + \omega_t \Delta t,
\end{aligned}
\end{equation}

where $\Delta t$ is the time step. 

\textbf{Complete Process} The entire process is summarized as follows:
\begin{itemize}
    \item \textbf{MP-SVES Optimization} refines learned guidance trajectory, ensuring it adheres to safety constraints while remaining as close as possible to learned guidance trajectory.
    \item \textbf{MPC Control} computes the optimal control inputs based on safety constrained trajectory, ensuring the robot follows the trajectory while respecting dynamic and physical constraints.
\end{itemize}
By combining MP-SVES optimization and MPC control, this approach enables robust trajectory tracking in dynamic and narrow environments.

\section{Experiments}
In this section, we conducted comprehensive experiments to evaluate the effectiveness of our method. Section~\ref{Experimental Setup} describes the experimental setup. Section~\ref{Guidance-Reuse under Environmental Perturbations} evaluates robustness when facing previously unseen obstacles. Section~\ref{Obstacle Avoidance without Guidance} assesses obstacle avoidance performance in unguided exploration within dense environments. Section~\ref{Obstacle Avoidance in Complex Environments} tests navigation in narrow corridors, requiring precise obstacle avoidance and path planning. Section~\ref{Real-world Experiments} validates our method in real-world scenarios, demonstrating safe and reliable navigation under practical conditions. Finally, Section~\ref{Ablation Study} analyzes the contribution of each module through ablation study.

\subsection{Experimental Setup}
\label{Experimental Setup}

\textbf{Baselines:} We compare our method against the following pure RGB visual navigation baselines:
\begin{itemize}
\item \textbf{General Navigation Model (GNM):} A general-purpose visual navigation policy trained on diverse datasets collected across multiple environments and robot platforms \cite{shah2022gnm}.
\item \textbf{Visual Navigation Transformer (ViNT):} A transformer-based foundation model for visual navigation that generates subgoals via diffusion-based planning \cite{shah2023vint}.
\item \textbf{Navigation with Goal Masked Diffusion (NoMaD):} A diffusion-based navigation policy that handles both goal-directed navigation and exploration in unknown environments \cite{sridhar2023nomad}.
\end{itemize}

\textbf{Datasets:} To ensure fair comparison, all methods are trained on the same set of navigation datasets:
\begin{itemize}
\item \textbf{SACSoN/HuRoN:} 70+ hours indoor trajectories with human interactions\cite{hirose2023sacson}.
\item \textbf{SCAND:} 9 hours of trajectories from Spot and Jackal robots in diverse social navigation scenarios \cite{karnan2022scand}.
\item \textbf{GoStanford2:} 16 hours of Turtlebot2 trajectories collected across multiple campus buildings \cite{hirose2018gonet}.
\item \textbf{RECON:} 18 months of navigation data supporting cross-domain tasks in varied environments \cite{shah2021rapid}.
\end{itemize}

\textbf{Training:} All methods are trained on RGB images and trajectory sequences from the datasets above, using a single NVIDIA RTX 4090 GPU, batch size 256, for 20 epochs.

\textbf{Evaluation Metrics:} We evaluate navigation performance using \textit{success rate} and \textit{average number of collisions per trial}. All simulation experiments and ablation study are repeated 50 times to ensure statistical reliability.

\textbf{Scenario Configuration:} All evaluation scenarios are unseen during training. The robot’s full trajectory is recorded within a fixed time horizon. A trial is considered successful if the robot reaches the designated goal region within this limit; otherwise, it is marked as a failure. Collisions are counted whenever the robot briefly contacts an obstacle or wall before reaching the goal.

\begin{figure}[t]
    \centering
    \vspace{5pt}
    \captionsetup{font=small}
    \includegraphics[width=\linewidth]{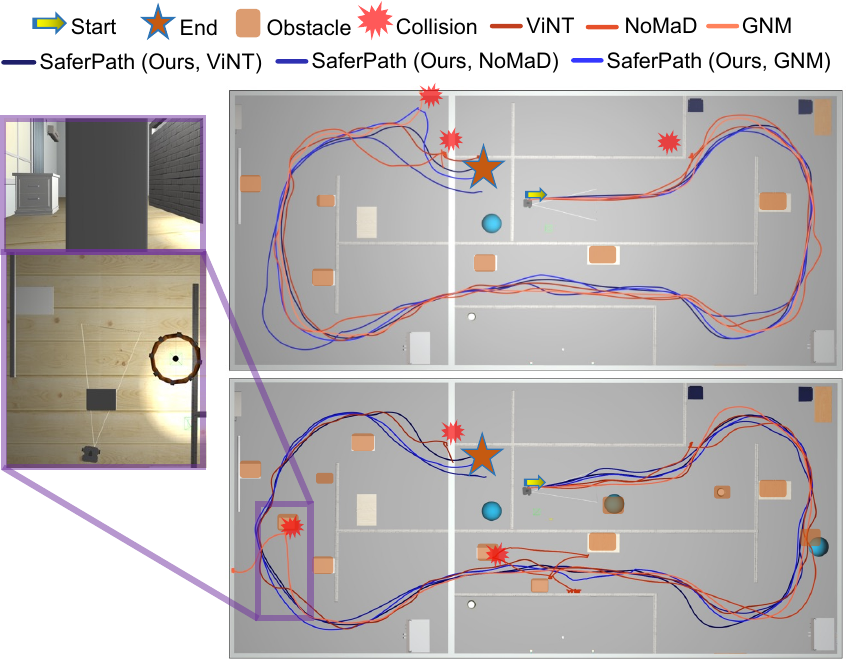}
    \caption{The figure illustrates the scene from Robust Navigation under Unseen Obstacles experiment.
The scene above (Scene A) introduces obstacles along the sides of the road in the original environment, while the scene below (Scene B) includes all the obstacles from Scene A and adds additional obstacles in the middle of the road. The figures display some examples of successful trajectories of our method and examples of collision scenarios of the baseline methods.}
    \label{fig:experiment1}
    \vspace{-15pt}
\end{figure}

\subsection{Robust Navigation under Unseen Obstacles}
\label{Guidance-Reuse under Environmental Perturbations}
In this experiment, we evaluate the robustness of our framework when the robot encounters obstacles that are absent from the learned guidance. Specifically, we use guidance generated in the original environment without these obstacles, and during deployment we introduce additional, previously unseen obstacles along the navigation path while keeping the guidance unchanged. This setup contains dynamic environmental changes and tests the model’s ability to handle unforeseen perturbations.

The experimental design is illustrated in Fig.~\ref{fig:experiment1}, where obstacles are randomly placed along indoor corridors to interfere with the planned navigation trajectory. In Scene A, obstacles are positioned along the sides of the corridors, causing minimal occlusion of the robot’s view. In Scene B, additional obstacles are placed in the middle of the corridors, creating significant occlusion and posing greater challenges to navigation. As shown in Fig.~\ref{fig:experiment1}, these occlusions can temporarily restrict the robot’s field of view, leading to partial or complete loss of guidance information and potentially unsafe trajectory predictions.



Table~\ref{tab:exp1_results} summarizes the results. SaferPath(Ours, GNM/ViNT/NoMaD) denotes our framework built on these end-to-end visual navigation models. In Scene A, with only a few additional obstacles, all baseline methods achieve moderate success rates. In Scene B, where occlusions are more severe, baseline performance drops significantly: GNM and NoMaD show larger declines, while ViNT remains relatively more robust. NoMaD also exhibits higher collision counts, likely due to its reliance on goal masks during training, which can lead to unsafe trajectories when guidance is unavailable; in contrast, GNM and ViNT produce more conservative behaviors under such conditions. When our module is integrated, all enhanced versions maintain high success rates and low collision counts across both Scene A and Scene B, with only minor performance reduction under heavy occlusion. These results demonstrate that our module effectively mitigates guidance unavailability, enabling safe and reliable navigation in dynamically changing environments.

\begin{table}[t]
\centering
\vspace{5pt}
\captionsetup{font=small}
\caption{Performance comparison of Robust Navigation under Unseen Obstacles experiment. Our method outperforms all baseline methods in the same scene. When more obstacles are introduced into the scene, our method experiences less performance drop compared to the baseline methods.}
\uniformtablesize
\resizebox{\linewidth}{!}{
\begin{tabular}{lcc c c}
\toprule
\textbf{Method} & \multicolumn{2}{c}{\textbf{Scene A}} & \multicolumn{2}{c}{\textbf{Scene B}} \\
\cmidrule(lr){2-3} \cmidrule(lr){4-5}
       & \textbf{Success (\%)} & \textbf{Coll.} & \textbf{Success (\%)} & \textbf{Coll.} \\
\midrule
GNM        & 68   & 0.70   & 26   & 1.22   \\
ViNT       & 94   & 0.22   & 80   & 0.80   \\
NoMaD      & 56   & 1.46   & 20   & 2.34   \\
\midrule
SaferPath(Ours, GNM)   & 82 & 0.08 & 74 & \textbf{0.06} \\
SaferPath(Ours, ViNT)  & 94 & \textbf{0}    & \textbf{94} & 0.08 \\
SaferPath(Ours, NoMaD) & \textbf{98} & 0.02 & \textbf{94} & \textbf{0.06} \\
\bottomrule
\end{tabular}}
\label{tab:exp1_results}
\end{table}

\begin{table}[t]
\centering
\captionsetup{font=small}
\caption{Performance comparison of Exploration in Dense Unstructured Environments experiment. Our method demonstrates significantly better obstacle avoidance capability compared to the baseline method.}
\label{tab:exp2_results}
\uniformtablesize
\begin{tabular}{lcc}
\toprule
\textbf{Method} & \textbf{Success (\%)} & \textbf{Coll.} \\
\midrule
NoMaD & 38 & 2.66 \\
SaferPath(Ours, NoMaD) & \textbf{96} & \textbf{0.04} \\
\bottomrule
\end{tabular}
\vspace{-15pt}
\end{table}

\subsection{Exploration in Dense Unstructured Environments}
\label{Obstacle Avoidance without Guidance}

\begin{figure}[t]
    \centering
    \vspace{5pt}
    \captionsetup{font=small}
    \includegraphics[width=0.9\linewidth]{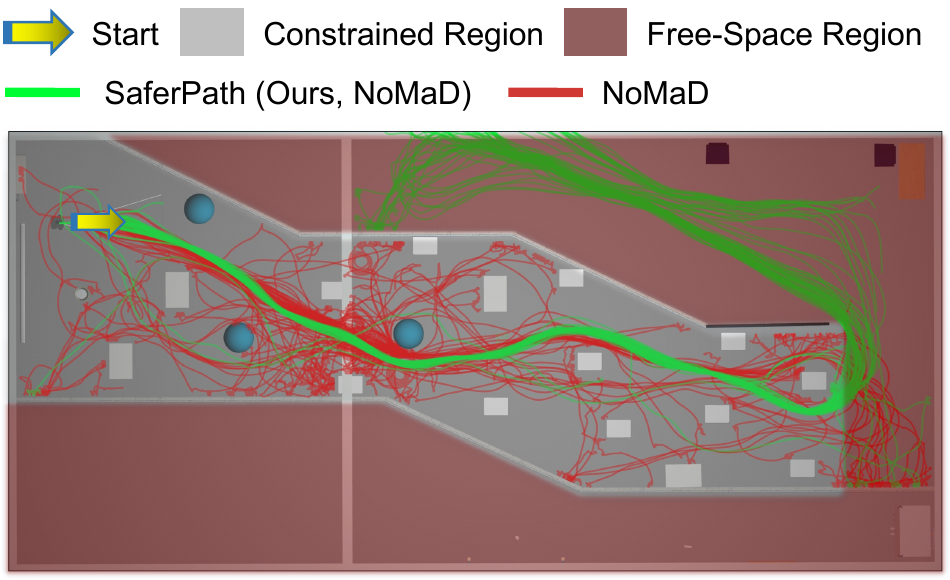}
    \caption{In Exploration in Dense Unstructured Environments experiment, the trajectory of SaferPath (Ours, NoMaD) (\textcolor{green}{green}) effectively avoids dense obstacles and reaches the free-space region. In contrast, the NoMaD trajectory (\textcolor{red}{red}) is more chaotic and results in significantly more collisions.}
    \label{fig:experiment2}
    \vspace{-5pt}
\end{figure}

In this experiment, we evaluate the framework’s ability to explore challenging unstructured environments with dense obstacles in the absence of explicit goal guidance. The method leverages its goal mask mechanism to enable exploration without external inputs. A dense obstacle field is constructed, and the robot is initialized within this region (see Fig.~\ref{fig:experiment2}). The task is considered successful if the robot exits the obstacle field within a predefined time limit. Performance is assessed using two metrics: success rate and average number of collisions.

Table~\ref{tab:exp2_results} reports the results. In dense obstacle fields without guidance, our method demonstrates effective obstacle avoidance and reliable navigation. Compared to NoMaD, it improves the success rate by over 50\% while also significantly reducing collisions, underscoring its robustness and safety in unguided exploration scenarios.

\subsection{Obstacle Avoidance in Narrow Corridors}
\label{Obstacle Avoidance in Complex Environments}

\begin{figure}[t]
    \centering
    \captionsetup{font=small}
    \includegraphics[width=\linewidth]{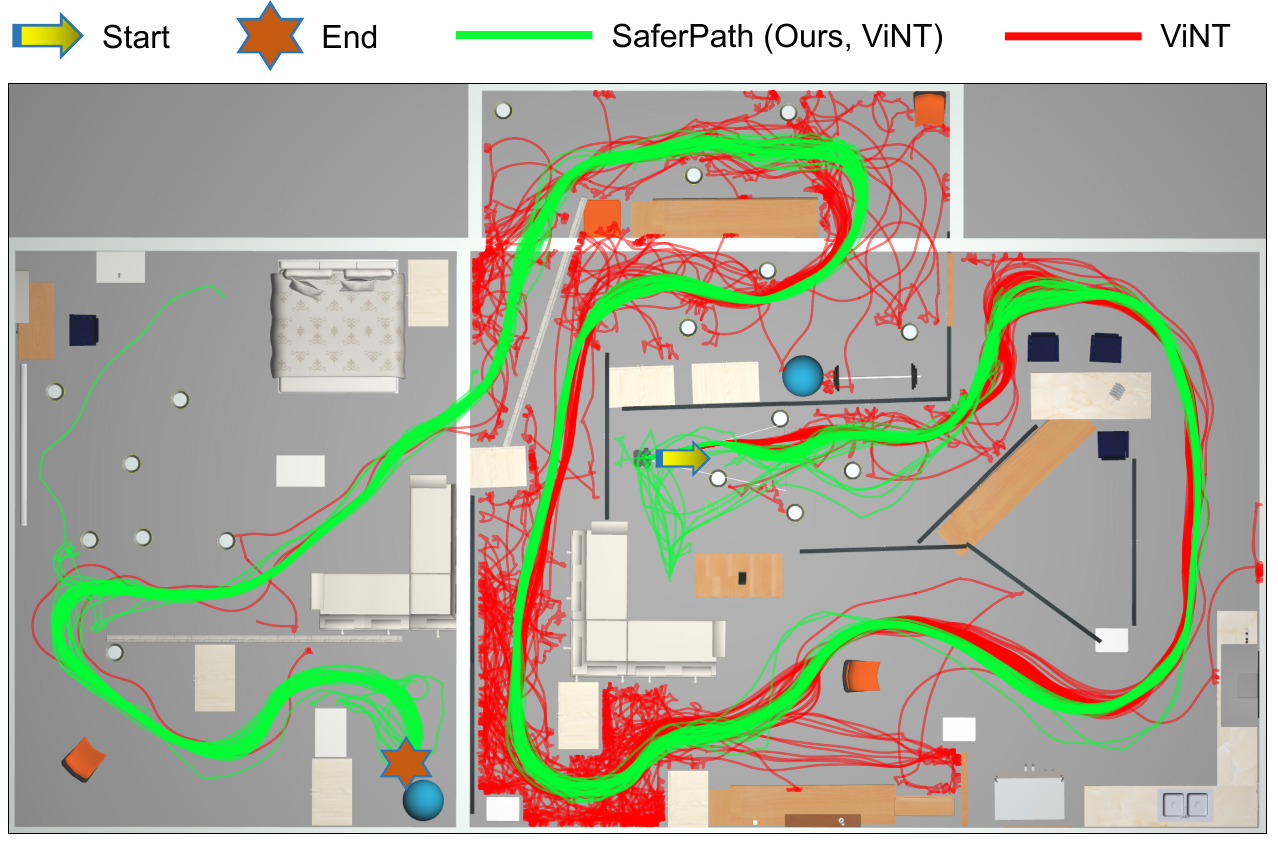}
    \caption{The figure illustrates partial trajectories from Obstacle Avoidance in Narrow Corridors experiment, where the trajectory of SaferPath (Ours, ViNT) (\textcolor{green}{green}) is clearly smoother and safer, with significantly fewer collisions compared to ViNT (\textcolor{red}{red}). This demonstrates the superior obstacle avoidance capability and robustness of SaferPath in complex environments over ViNT.}
    \label{fig:experiment3}
    \vspace{-15pt}
\end{figure}

In this experiment, we evaluate navigation performance in narrow and unstructured indoor environments (see Fig.~\ref{fig:experiment3}). This setting imposes stringent requirements on accurate obstacle avoidance and precise path planning through confined corridors, where even small deviations can result in wall collisions. The irregular layout of the passageways further challenges the real-time performance and robustness of different methods.

Table~\ref{tab:exp3_results} summarizes the results. All baseline methods fail in this demanding navigation task, unable to maintain safe trajectories in such constrained spaces. In contrast, the three enhanced versions equipped with our module achieve significantly higher success rates and lower collision counts. Taking ViNT as an example, we visualize the trajectories from all trials for both the original and enhanced versions. The results clearly show that our module enables the robot to remain within safe passage areas, while the original ViNT frequently produces unsafe trajectories that lead to collisions and task failure.

\begin{figure*}[t]
    \centering
    \vspace{5pt}
    \captionsetup{font=small}
    \includegraphics[width=0.78\linewidth]{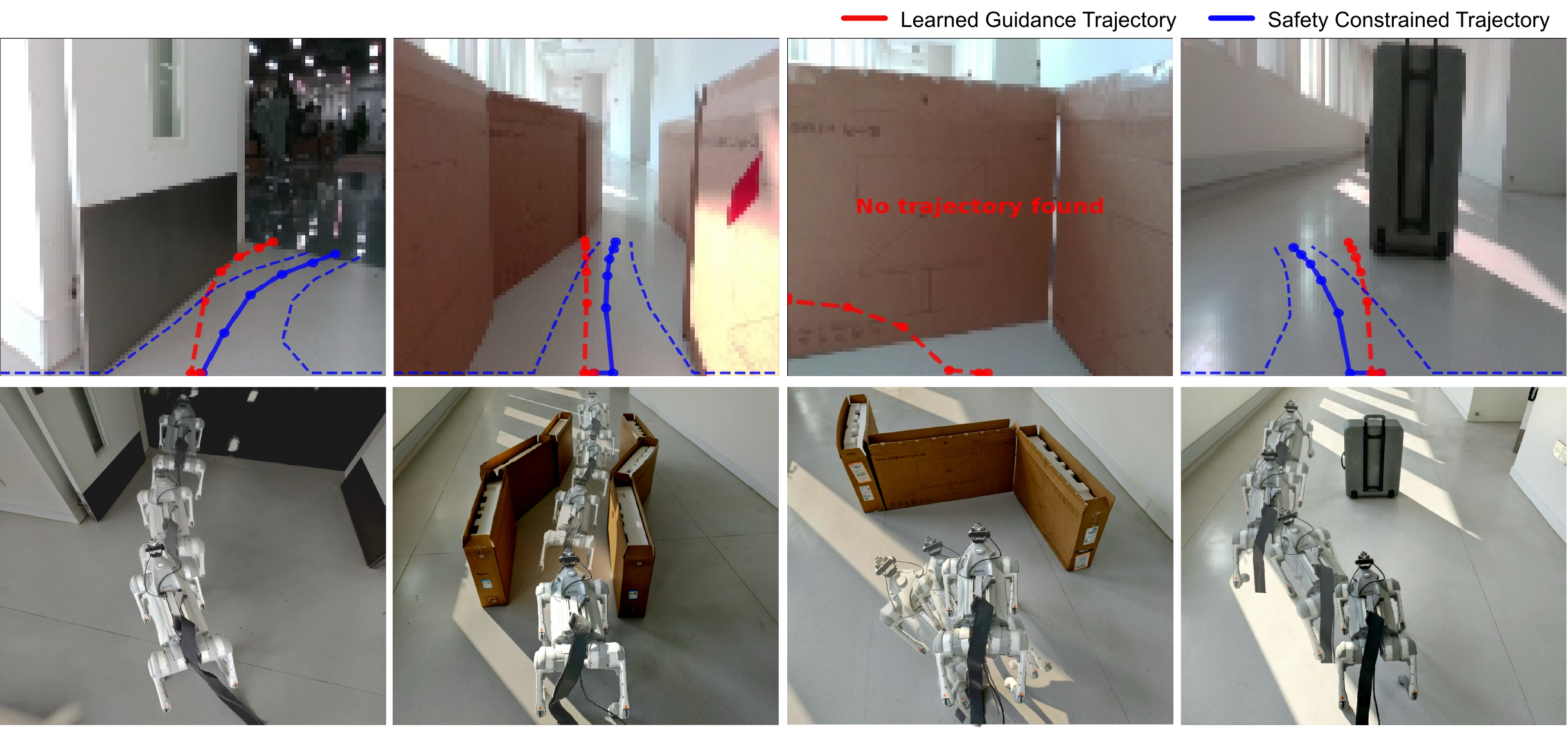}
    \caption{Our SaferPath deployed on the Unitree Go2 successfully avoids both unseen and dynamic obstacles, navigates narrow spaces, and consistently remains within safe areas. When no traversable trajectory is available, the Emergency Indicator stops the robot and reorients it toward a safe direction.}
    \label{fig:real_experiment}
    \vspace{-15pt}
\end{figure*}

\begin{table}[t]
\centering
\captionsetup{font=small}
\caption{Performance comparison in Obstacle Avoidance in Narrow Corridors experiment. Our method significantly outperforms the baseline methods, with all baseline methods failing in this experiment.}
\label{tab:exp3_results}
\uniformtablesize
\begin{tabular}{lcc}
\toprule
\textbf{Method} & \textbf{Success (\%)} & \textbf{Coll.} \\
\midrule
GNM & 0 & 4.36 \\
NoMaD & 0 & 4.10 \\
ViNT & 0 & 4.04 \\
\midrule
SaferPath(Ours, GNM) & 76 & 0.42 \\
SaferPath(Ours, NoMaD) & 54 & 0.52 \\
SaferPath(Ours, ViNT) & \textbf{80} & \textbf{0.16} \\
\bottomrule
\end{tabular}
\vspace{-15pt}
\end{table}

\subsection{Real-World Navigation on a Quadruped Robot}
\label{Real-world Experiments}

We conducted real-world experiments to validate the effectiveness of our method in practical scenarios. The experiments were performed on a Unitree Go2 quadruped robot equipped with an Intel D435i camera, using only its RGB functionality for perception. For navigation, we deployed SaferPath (ours) alongside NoMaD for comparison.
Figure~\ref{fig:real_experiment} shows representative scenarios where four baseline methods are prone to collisions. In these cases, the reference trajectories often bring the robot dangerously close to obstacles. By contrast, after integrating SaferPath, the robot is able to (i) maintain a safe distance from side walls, (ii) navigate narrow passages without collisions, (iii) react promptly to emergency situations by stopping and rerouting to a feasible path, and (iv) avoid obstacles positioned in the center of its trajectory. These results demonstrate that SaferPath enables robust and safe navigation in real-world environments, effectively handling narrow passages and dynamic obstacles.

\subsection{Ablation Study}
\label{Ablation Study}
To analyze the contribution of individual components, we conduct an ablation study by selectively removing the Traversability Score Mapper (TSM), MP-SVES, and Emergency Indicator modules:

\begin{itemize}
    \item \textit{w/o TSM}: Remove the Traversability Score Mapper, eliminating the score-map safety constraint.
    \item \textit{w/o MP-SVES}: Remove the MP-SVES module, placing all constraints into the MPC module and solving with a conventional nonlinear solver IPOPT\cite{ipopt}.
    \item \textit{w/o Em}: Remove the Emergency Indicator module, such that when a safety-constrained trajectory is infeasible, the learned guidance trajectory is executed directly.
\end{itemize}


We repeat the obstacle avoidance in complex environments experiment using ViNT as the underlying end-to-end visual navigation model. Table~\ref{tab:exp4_results} summarizes the results. Removing any module degrades performance. In particular, eliminating MP-SVES causes the success rate to drop to 0\%, underscoring its critical role in ensuring system stability. Without the Traversability Score Mapper, the success rate decreases by 20\% and collisions increase significantly, highlighting its importance for obstacle avoidance. Removing the Emergency Indicator module also reduces success rates and increases collisions, showing its necessity in extreme cases.

\begin{table}[t]
\centering
\captionsetup{font=small}
\caption{The ablation study reveals that all modules are essential, with MP-SVES being the most critical for system stability and performance.}
\label{tab:exp4_results}
\uniformtablesize
\begin{tabular}{lcc}
\toprule
\textbf{Method} & \textbf{Success (\%)} & \textbf{Coll.} \\
\midrule
SaferPath(Ours w/o TSM, ViNT) & 60 & 1.34 \\
SaferPath(Ours w/o MP-SVES, ViNT) & 0  & 2.28 \\
SaferPath(Ours w/o Em, ViNT) & 70 & 1.14 \\
SaferPath(Ours, ViNT) & \textbf{80} & \textbf{0.16} \\
\bottomrule
\end{tabular}
\vspace{-10pt}
\end{table}

These results directly address the research questions posed in the Introduction. For \textbf{Q1}, the proposed module significantly improves the safety and robustness of end-to-end visual navigation: in environments with additional obstacles, augmented models maintain high success rates and low collision counts, demonstrating that feasible and safe trajectories are preserved. For \textbf{Q2}, in narrow and cluttered environments that are underrepresented in training data, the module enables efficient and collision-free navigation, yielding substantial performance gains. Consistent results across both simulation and real-world experiments confirm the effectiveness and generality of our approach.
\section{Conclusion}

We propose SaferPath, a hierarchical visual navigation framework that leverages learned guidance from end-to-end visual navigation models while ensuring safe and feasible trajectory execution. By effectively utilizing the directional information provided by these models, SaferPath employs the MP-SVES algorithm to optimize trajectories, generating safety constrained paths in just a few iterations while handling both safe zone navigation and emergency obstacle avoidance. Extensive experiments demonstrate that our approach outperforms state-of-the-art end-to-end visual models by over 40\%, and is capable of safe navigation even in complex and narrow environments.

Despite these improvements, the method is limited by the restricted field of view of RGB observations, which may lead to collisions when obstacles are outside the camera's view. Future work will focus on incorporating richer sensory modalities and predictive scene understanding to further improve robustness in cluttered and dynamic settings.

\bibliographystyle{IEEEtran}
\bibliography{reference}

\end{document}